# POLYGLOT-NER: Massive Multilingual Named Entity Recognition

Rami Al-Rfou Vivek Kulkarni Bryan Perozzi
Steven Skiena
Department of Computer Science
Stony Brook University
{ralrfou, vvkulkarni, bperozzi, skiena}@cs.stonybrook.edu

**Abstract**

The increasing diversity of languages used on the web introduces a new level of complexity to Information Retrieval (IR) systems. We can no longer assume that textual content is written in one language or even the same language family.

In this paper, we demonstrate how to build massive multilingual annotators with minimal human expertise and intervention. We describe a system that builds Named Entity Recognition (NER) annotators for **40 major languages** using Wikipedia and Freebase. Our approach does not require NER human annotated datasets or language specific resources like treebanks, parallel corpora, and orthographic rules. The novelty of approach lies therein - using only language agnostic techniques, while achieving competitive performance.

Our method learns distributed word representations (word embeddings) which encode semantic and syntactic features of words in each language. Then, we automatically generate datasets from Wikipedia link structure and Freebase attributes. Finally, we apply two preprocessing stages (oversampling and exact surface form matching) which do not require any linguistic expertise.

Our evaluation is two fold: First, we demonstrate the system performance on human annotated datasets. Second, for languages where no gold-standard benchmarks are available, we propose a new method, *distant evaluation*, based on statistical machine translation.

## 1 Introduction

The growth of the Internet is bringing new communities, cultures and languages online. However, not enough work has been proposed to deal with the increasing linguistic variety of web content. Natural Language Processing (NLP) tools are limited to a small number of languages, usually only English, which does not reflect the fast changing pace of the Internet. Correspondingly, current multilingual text-based Information Retrieval (IR) systems are restricted to simple processing stages that are based on word's surface forms and frequency based methods. We believe addressing the multilingual aspect of the problem is crucial for the future success of these systems.

In this work, we perform a case study on how to build a massively multilingual Named Entity Recognition (NER) system. The Named Entity Recognition task (also known as *entity extraction* or *entity identification*) extracts chunks of text as phrases and classifies them into pre-defined categories such as the names of persons, locations, and organizations. NER is an essential pre-processing stage in NLP and Information Retrieval (IR) systems, where it is used for a variety of purposes (e.g, event extraction or knowledge base population). Successful approaches to address NER rely on supervised learning [5, 12]. Applying these approaches to a massively multilingual setting exposes two major drawbacks; First, they require human annotated datasets which are scarce. Second, to design relevant features, sufficient linguistic proficiency is required for each language of interest. This makes building multilingual NER annotators a tedious and cumbersome process.

Our work addresses these drawbacks by relying on language-independent techniques. We use neural word embeddings, Wikipedia link structure, and Freebase attributes to automatically construct NER annotators for 40 major languages. First, we learn neural word embeddings which encode semantic and syntactic features of words in each language. Second, we use the internal links embedded in Wikipedia articles to detect named entity mentions. When a link points to an article identified by Freebase as an entity article, we include the anchor text as a positive training example. However, not all entity mentions are linked in Wikipedia because of style guidelines. To address this problem, we propose oversampling and surface word matching to solve this *positive-only label learning* problem [10, 16–18] while avoiding any language-specific dependencies.

Lack of human annotated datasets not only limits quality of training but also system evaluation. We evaluate on standard NER datasets if they are available. For the remaining languages, we propose *distant evaluation* based on statistical machine translation (SMT) to generate testing datasets that provide insightful analysis of the system performance.
In summary, our contributions are the following:

- **Language-independent extraction** - for noisy datasets. Our proposed *language-agnostic* techniques address noise introduced by Wikipedia style guidelines, boosting the performance by at least $45\%$ $F_1$ on human annotated gold standards.
- **40 NER annotators**[1] - We are releasing the trained

[1] Online demo is available at `https://bit.ly/polyglot-ner`.

| | Preprocessing stages | | | | | | | Languages Covered | |
|---|---|---|---|---|---|---|---|---|---|
| | *Language Specific* | | | | | *Agnostic* | | | |
| | $S_1$ | $S_2$ | $S_3$ | $S_4$ | $S_5$ | $S_6$ | $S_7$ | # | Language code |
| Toral and Munoz [26] | | | ✓ | | ✓ | ✓ | | 1 | *en* |
| Kazama and Torisawa [14] | | ✓ | ✓ | | | | | 1 | *en* |
| Richman and Schone [23] | ✓ | ✓ | | | | ✓ | | 7 | *en, es, fr, uk, ru, pl, pt* |
| Ehrmann et al. [9] | ✓ | ✓ | | ✓ | | ✓ | | 6 | *en, es, fr, de, ru, cs* |
| Kim et al. [15] | ✓ | | | ✓ | | ✓ | | 3 | *en, ko, bg* |
| Nothman et al. [22] | ✓ | ✓ | ✓ | | | ✓ | | 9 | *en, es, fr, de, ru, pl, pt, it, nl* |
| POLYGLOT-NER | | | | | | ✓ | ✓ | **40** | *en, es, fr, de, ru, pl, pt, it, nl ar, he, hi, zh, ko, ja, tl, ms, . . .* |

| | |
|---|---|
| $S_1$ | Orthographic features. |
| $S_2$ | Gazetteers and Dictionaries. |
| $S_3$ | Part of speech tags. |
| $S_4$ | Parallel corpora and projected annotations. |
| $S_5$ | Wordnet. |
| $S_6$ | Exact surface form matching. |
| $S_7$ | Oversampling. |

Table 1: Proposed systems preprocessing stages and language coverage. Processing stages are divided into two groups, techniques that require *language specific* dependencies ($S_1 - S_5$) and others that can be applied to any language without restrictions ($S_6, S_7$). The novelty lies in using language agnostic techniques, yet achieving comparable performance.

models as open source software. These annotators are invaluable, especially for resource scarce languages, like Serbian, Indonesian, Thai, Malay and Hebrew.

- **Distant Evaluation** - We propose a technique based on statistical machine translation to scale our evaluation in the absence of human annotated datasets.

Our paper is structured as follows: First, we review the related work in Section 2. In Section 3, we present our formulation of the NER problem and describe our semi-supervised approach to build annotators (models) for 40 languages. Section 4 shows our procedure to generate training datasets using Wikipedia and Freebase. We discuss our results in Section 5. Section 6 shows how statistical machine translation is used to evaluate the performance of our system.

## 2 Related Work

Wikipedia has been used as a resource for many tasks in NLP and IR [13, 19]. There is a body of literature regarding preprocessing Wikipedia for NER [9, 14, 15, 22, 23, 26] which is summarized in Table 1. All previous work depends on language specific preprocessing stages such as as taggers and parallel corpora. The reliance on language specific processing poses a bottleneck to the scalability and diversity of the languages covered by the previous systems. In contrast, our work relies on only language agnostic techniques.

The closest related work is Nothman et al. [22]. Compared to their approach, we find that using only oversampling is a sufficient replacement for their entire proposed language dependent preprocessing pipeline (See Section 4.2.1).

## 3 Semi-supervised Learning

The goal of Named Entity Recognition (NER) is to identify sequences of tokens which are entities, and classify them into one of several categories. We follow the approach proposed by [7] to model NER as a word level classification problem. They observe that for most chunking tasks, including NER, the tag of a word depends mainly on its neighboring words. Considering only local context yields models with competitive performance to schemes which take into account the whole sentence structure. This word level approach ignores the dependencies between word tags and thus might not capture some constraints on tags' appearance order in the text. However, empirically, our evaluation does not indicate the manifestation of this problem. More importantly, this word based formulation allows us to use simpler oversampling and exact-matching mechanisms as we will see in Section 4.

**3.1 Word Embeddings** capture semantic and syntactic characteristics of words through unsupervised learning [20]. They have been successfully used as features for several tasks including NER [7, 27] and proposed as a cornerstone for developing multilingual applications [1].

Word embeddings are latent representations of words acquired by harnessing huge amounts of raw text through language modeling. These representations capture information about word co-occurrences and therefore their syntactic functionality and semantics. Given the abundance of unstructured text available online, we can automatically learn these embeddings for all languages and use them as features in an *unsupervised* manner.

More specifically, given a language with vocabulary $V$, a word embedding is a mapping function $\Phi\colon w \mapsto \mathbb{R}^d$, where $w \in V$ and $d$ is a constant value that ranges usually between 50 and 500. We use the Polyglot embeddings [1] as our sole features for each language under investigation[2]. The Polyglot embeddings are trained on Wikipedia without any labelled data, the vocabulary of each language consists of the most frequent 100K words and the word representation consist of 64 dimensions ($d = 64$). The Polyglot embeddings were trained using an objective function proposed by [6] which takes *ordered* sequences of words as its input. Therefore, the learned representations cluster words according to their part of speech tags. Given that most of named entities are proper nouns (part of speech), these representations are a natural fit to our task.

**3.2 Discriminative Learning** We model NER as a word level classification problem. More formally, let $\mathcal{W}_i^n = (w_{i-n} \cdots w_i \cdots w_{i+n})$ be a phrase centered around the word $w_i$ with a window of size $2n+1$. We seek to learn a target function $F\colon \mathcal{W}_i^n \mapsto \mathcal{Y}$, where $\mathcal{Y}$ is the set of tags. First, we map the phrase $\mathcal{W}_i^n$ to its embedding representation

$$\mathbf{\Phi}_i^n = [\Phi(w_{i-n}); \ldots; \Phi(w_i); \ldots; \Phi(w_{i+n})].$$

Next, we learn a model $\Psi_y$ to score tag $y$ given $\mathbf{\Phi}_i^n$, i.e $\Psi_y : \mathbf{\Phi}_i^n \mapsto \mathbb{R}$, using a neural network with one hidden layer of size $h$

$$\Psi_y(\mathbf{\Phi}_i^n) = \mathbf{s}^T(\tanh(\mathbf{W}\mathbf{\Phi}_i^n + \mathbf{b})), \tag{3.1}$$

where $\mathbf{W} \in \mathbb{R}^{h(2n+1)d}$ and $\mathbf{s} \in \mathbb{R}^h$ are the first and second layer weights of the neural network, and $\mathbf{b} \in \mathbb{R}^h$ are the bias units of the hidden layer. Finally, we construct a one-vs-all classifier $F$ and penalize it by the following hinge loss,

$$J = \frac{1}{m}\sum_{i=1}^{m} max\Big(0,\ 1 - \Psi_{t_i}(\mathbf{\Phi}_i^n) + \max_{\substack{y \neq t_i \\ y \in \mathcal{Y}}} \Psi_y(\mathbf{\Phi}_i^n)\Big),$$

where $t_i$ is the correct tag of the word $w_i$, and $m$ is the size of the training set.

**3.3 Optimization** We learn the parameters $\theta = (\mathbf{s}, \mathbf{W}, \mathbf{b})$ via backpropagation with stochastic gradient descent. As the stochastic optimization performance is dependent on a good choice of the learning rate, we automate the learning rate selection through an adaptive update procedure [8]. This results in separate learning rates $\eta_i$ for each individual parameter, $\theta_i$. More specifically the learning rate at step $t$ for parameter $i$ is given by the following:

$$\eta_i(t) = \frac{1.0}{\sqrt{\sum_{s=1}^{t}\left(\frac{\partial J(s)}{\partial \theta_i(s)}\right)^2}}. \tag{3.2}$$

For the rest of paper, we train our annotators for 50 epochs with the adaptive learning values uniformly initialized with mean equal to zero.

To understand the upper bound of performance we can achieve through this specific modeling of NER. We trained new word embeddings with extended vocabulary (300K words) using English, Spanish and Dutch Wikipedia. Table 2 shows the performance of our annotators given CoNLL training datasets [24, 25] and the word embeddings as features. Our results on English are similar to the ones reported by [28]. [3]

| | Dev | Test |
|---|---|---|
| English (Wang et. al, 2013) | 86.6 | 80.7 |
| English (Ours) | 88.9 | 84.8 |
| Spanish (Ours) | 70.2 | 70.8 |
| Dutch (Ours) | 67.8 | 69.2 |

Table 2: Exact $F_1$ scores of our annotators trained and tested on CoNLL.

## 4 Extracting Entity Mentions

Our procedure for creating a named entity training corpus from Wikipedia consists of two steps; First, we find which Wikipedia articles correspond to entities, using Freebase [4]. Second, we cope with the missing annotations problem through oversampling from the entity classes and extending the annotation coverage using an exact surface form matching rule.

**4.1 Article Categorization** Freebase maintains several attributes for each Wikipedia article, covering around 40 different Wikipedia languages. We categorize the article topics into one of the following categories, $\mathcal{Y}$ = {PERSON, LOCATION, ORGANIZATION, NONENTITY}. For each entity category we specify the corresponding freebase attributes, as the following:

| | |
|---|---|
| • LOCATION | /location/{citytown, country, region, continent, neighborhood, administrative_division} |
| • ORGANIZATION | /sports/sports_team, /book/newspaper, /organization/organization, |
| • PERSON | /people/person |

The result is a mapping of Wikipedia page titles and their redirects to $\mathcal{Y}$. If an internal link points to any of these titles, we consider it an entity mention. Table 3 shows the percentage of pages that are covered by Freebase for some of the languages we consider. The entity coverage varies with each language, and this greatly biases the label distribution of the generated training data. We will overcome this bias in the distribution of entity examples using oversampling (See Section 4.2.1).

[2] Available: http://bit.ly/embeddings

[3] Notice, that we obtain higher results, here, than annotators trained on Wikipedia datasets (Table 6), because training and testing datasets belong to the same domain.

| Language | Coverage | Language | Coverage |
|---|---|---|---|
| Malay | 37.8% | Arabic | 15.6% |
| English | 35.2% | Dutch | 14.7% |
| Spanish | 24.8% | Swedish | 10.2% |
| Greek | 24.3% | Hindi | 8.8% |

Table 3: Percentage of Wikipedia articles identified by Freebase as entity based articles. There is a wide disparity across languages.

**4.2 Missing Links** Unfortunately, generating a training dataset directly from the link structure results in very poor performance with DEV$_{F_1}$ $< 10\%$ in English, Spanish and Dutch due to missing annotations. This is a consequence of Wikipedia style guidelines[4]. Editors are instructed to link the first mention in the page, but not later ones. This results in leaving most entity mentions unmarked. Table 4 examines this effect by contrasting the percentage of words that are covered by entity phrases in both CONLL and Wikipedia.

| | CONLL | WIKI |
|---|---|---|
| English | 16.76% | 2.34% |
| Spanish | 12.60% | 2.12% |
| Dutch | 9.29% | 2.28% |

Table 4: Percentage of words that are covered by entity phrases in each corpus.

We can view the generated examples as two sets; one that is truly positive and the other as a mix of negative and positive examples. Our task is to learn an annotator only from the positive examples. In such a setting, [10, 17, 18] show that considering the unlabeled set as negative examples while modifying the objective loss to accommodate different penalties for misclassifying each set of examples outperforms other heuristics and other iterative EM-based methods. Changing the label distribution by oversampling the positive labels will achieve a similar effect.

**4.2.1 Oversampling** To overcome the effect of missing annotations, we *correct* the label distribution by oversampling from the entity classes. The intuition is that untagged words are not necessarily non-entities. Conversely, we have high confidence in the links which have been explicitly tagged by users. To reflect the difference in confidence levels, we categorize our labels into two subcategories: $\mathcal{Y}^+ = \{$PERSON, LOCATION, ORGANIZATION$\}$ and $\mathcal{Y}^- = \{$NONENTITY$\}$. A training example $\mathbf{\Phi}_i^n$ is considered positive if $F(\mathbf{\Phi}_i^n) \in \mathcal{Y}^+$ and negative otherwise. We define the oversampling ratio ($\rho$) to be

$$\rho = \frac{\sum_i^m [F(\mathbf{\Phi}_i^n) \in \mathcal{Y}^+]}{m},$$

where $[x]$ is the indicator function and $m$ is the total number of training examples.

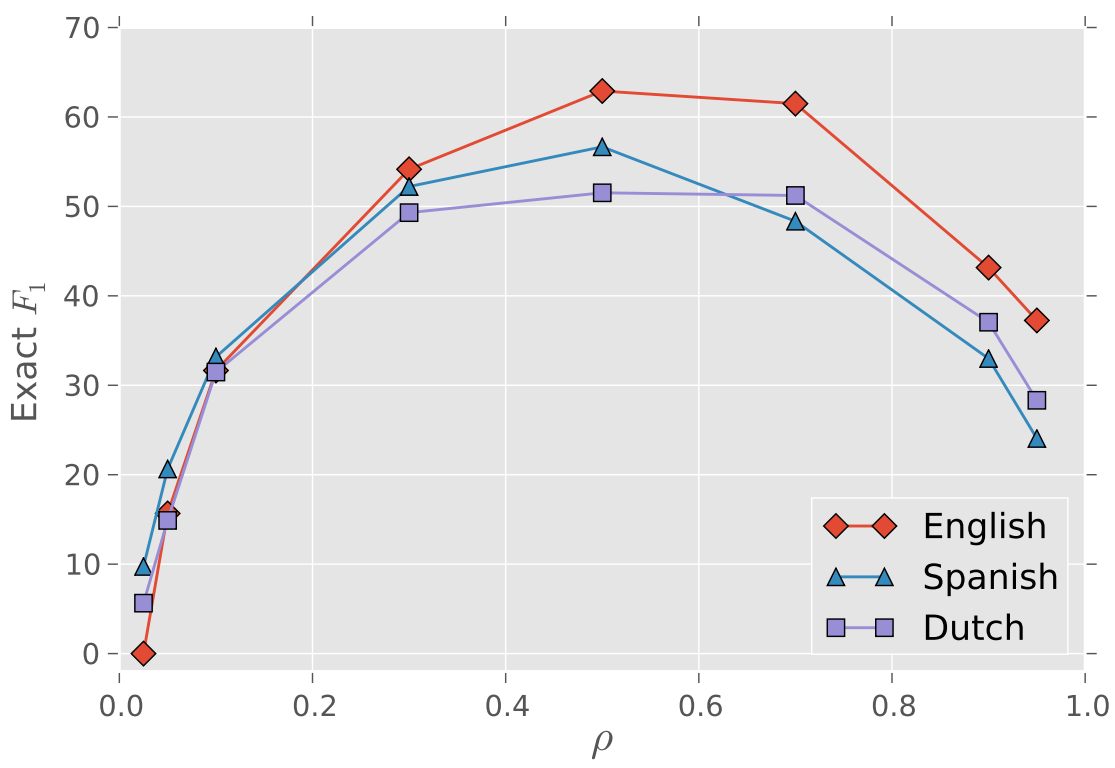

Figure 1: Performance of oversampling entities (POLYGLOT-NER$_{S_7}$) on DEV datasets. $\rho \cong 2.5\%$ corresponds to the original distribution of positive labels in Wikipedia text.

Our goal, here, is to construct a subset of our training corpus where $\rho$ is higher in this subset than the original training dataset. We sample the positive class uniformly without replacement. This insures we do not change the conditional distribution of a specific entity class given it is a positive example.

Figure 1 shows the effect of oversampling. The first point corresponds to the original distribution of positive labels in Wikipedia text, where $\rho \cong 2.5\%$. We observe that regardless of the chosen $\rho$, oversampling improves the results. This improvement is quite stable when $0.25 < \rho < 0.75$, and the Exact $F_1$ score is increased by at least 40% for all languages we consider. We choose $\rho = 0.5$ to be the value we use for our testing phase in Section 5 as it produces the maximum results across the three languages under investigation.

**4.2.2 Exact Surface Form Matching** While oversampling mitigates the effect of the skewed label distribution, it does not address the stylistic bias with which Wikipedia editors create links. The first bias is to link only the first mention of a entity in an article. This canonical mention is usually the *full* name of an entity, and not the abbreviated form used throughout the remainder of the article. We found that in 200K examples tagged with PERSON, 45K examples belong to three terms mentions, 140K to two terms mentions and only 15K belonging to single term mentions. This bias against single term mentions in links does not reflect the true distribution of named entities in the text resulting in annotators which tag `Noam Chomsky` but not `Chomsky`. The second bias is that there are no self-referential links inside an entity's article (e.g. on `Barack Obama`'s page, none of the sentences mentioning him are linked).

[4] http://en.wikipedia.org/wiki/Wikipedia:Manual_of_Style/Linking

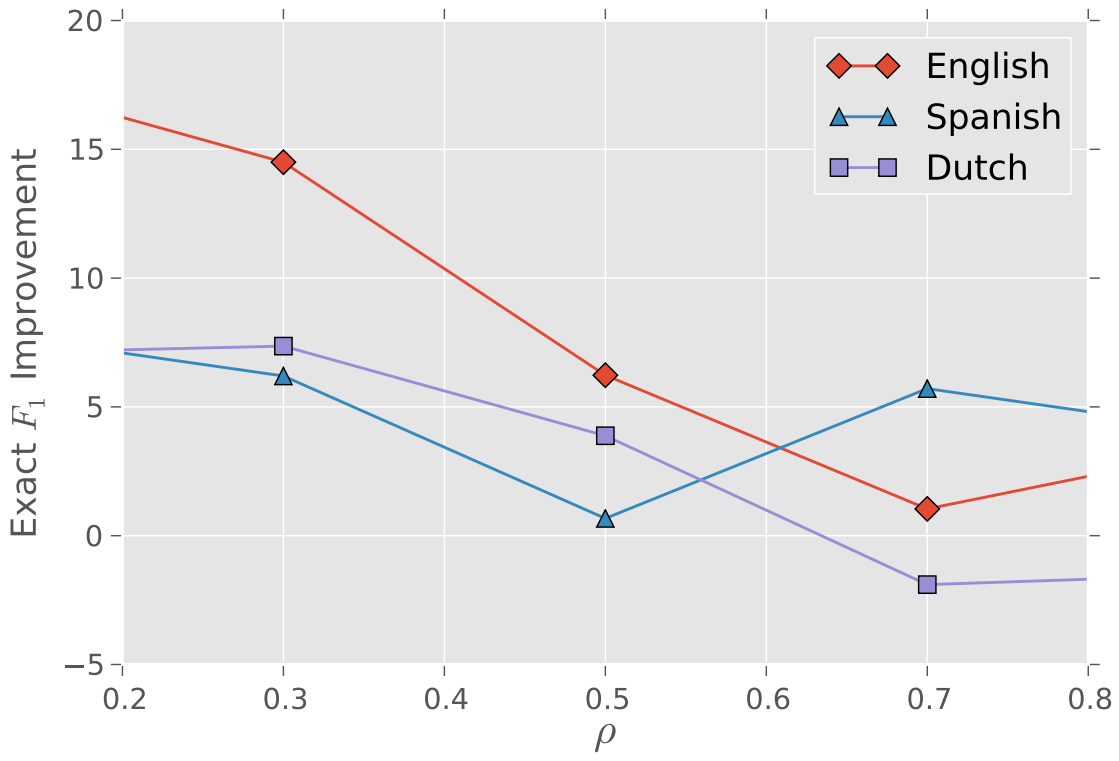

Figure 2: Performance improvement on CoNLL DEV datasets after applying coreference stage first on Wikipedia before oversampling.

In order to extend our annotations coverage on each article, we apply a simple rule of surface form matching. If a word appeared in an entity mention, or in the title of the current article (to address the second bias), we consider all appearances of this word in the current article to be annotated with the same tag. In the case of having multiple tags for the same word, we use the most frequent tag. This process can be viewed as a first-order coreference resolution where we link mentions using exact string matching.

For example, after this procedure, every mention of '`Barack Obama`', `Barack`, and `Obama` in the article on `Barack_Obama` will be considered a link referring to a PERSON entity. In order to avoid mislabeling functional words which appear in links (e.g. `of`, `the`, `de`) we exclude the most frequent 1000 words in our vocabulary.

Figure 2 shows the improvement this stage adds when it is applied to Wikipedia before oversampling. $F_1$ improvements that we observe in English, Spanish, and Dutch are significant, especially when $\rho \leq 0.5$. Most of this improvement is due to higher recall on the tag PERSON.

## 5 Results

In this section, we will analyze the errors produced system for qualitative assessment. In addition, we evaluate the performance of POLYGLOT-NER on CoNLL datasets to demonstrate the efficiency of the proposed solutions to deal with missing links in the Wikipedia markup.

Table 5 shows annotated examples for 11 different languages. Table 5a shows correctly annotated examples and Table 5b a sample of the mistakes our annotators make. Analyzing our good examples shows that our system performs the best on the PERSON category, even for names that are transliterated from other languages (see Russian, Arabic and Korean examples). Moreover, our system is still able to identify entities in mixed languages scenario, for example, the appearance of '`Spain`' in the Arabic example and '`Pešek`' in the Spanish example. This robustness stems from two factors; First, the embeddings vocabulary of a specific language includes frequent foreign words. Second, our annotators are able to capture sufficient local contextual clues.

| DEV | English | Spanish | Dutch |
|---|---|---|---|
| POLYGLOT-NER$_{S_7}$ | 62.9 | 56.7 | 53.2 |
| POLYGLOT-NER$_{S_6+S_7}$ | **73.3** | 59.3 | 59.7 |
| Nothman et al. [22] | 67.9 | **60.7** | **62.2** |
| TEST | | | |
| POLYGLOT-NER$_{S_7}$ | 58.5 | 58.5 | 51.5 |
| POLYGLOT-NER$_{S_6+S_7}$ | **71.3** | **63.0** | 59.6 |
| Nothman et al. [22] | 61.3 | 61.0 | **64.0** |

Table 6: Cross-domain performance measured by Exact $F_1$ on TEST and DEV sections of CoNLL corpora.

Our errors can be grouped into three categories (See Table 5b); First, common words {`River`, `House`} that appear in organizations are hard to identify. Second, our system does not consistently tag demonyms (nationalities), as Russian, French, Arabic, Greek examples show. Misclassification errors occur, common cases include confusion between LOCATION and ORGANIZATION tags in the case of nested entities (See the Chinese example) or between the PERSON and ORGANIZATION tags when company names are referred to in the same context as that of persons (See the Spanish example).

In addition to the qualitative analysis, we evaluate our models quantitatively on the CoNLL 2002 Spanish and Dutch datasets, and the CoNLL 2003 English dataset. We show results of our models trained on Wikipedia and evaluated on CoNLL in Table 6. Observe that oversampling (POLYGLOT-NER$_{S_7}$) alone is able to get competitive results. With exact surface form matching applied to the data first (POLYGLOT-NER$_{S_6+S_7}$), we outperform previous work on English and Spanish without applying any language-specific rules. Most of POLYGLOT-NER Dutch errors appear in the category of ORGANIZATION.

Training on Wikipedia results in lower scores on CoNLL testing datasets compared to models trained on CoNLL directly (See Table 2) due to two main factors. First this is an *out-of-domain* evaluation. Second, there are a variety of orthographic and contextual differences between Wikipedia and CoNLL. Common differences between the datasets include: trailing periods, leading delimiters, and modifiers

| Language | Sentence | Translation |
|---|---|---|
| English | **Simien** was traded from the Heat along with **Antoine Walker** and **Michael Doleac** to the **Minnesota Timberwolves** on October 24, 2007, for **Ricky Davis** and **Mark Blount**. | - |
| Hungarian | **Dimitri** beszélt egy utat **Rómába**. | **Dimitri** talked about a trip to **Rome**. |
| Spanish | **Pešek** nació en **Praga** y estudió dirección de orquesta piano en la **Academia de Artes** allí, con **Václav Smetacek** | **Pešek** born in **Prague** and studied orchestra direction piano at the **Academy of Arts** there, **Vaclav Smetacek**. |
| Russian | Уроженец Рио-де-Жанейро, Хосе Родригес Триндади использовал сокращенную форму как своим сценическим псевдонимом. | A native of **Rio de Janeiro**, **Jose Rodriguez Trindade** used as a shortened form of his stage name. |
| Korean | 도널드 스미스 는 1976 년에 자이드 압둘 아지즈 에 그의 이름 을 바꿨다 . | **Donald Smith** in 1976 changed his name to **Zaid Abdul Aziz**. |
| French | La **France** veut satisfaire à ses engagements envers l'**Union européenne**. | **France** wants to meet its commitments to the **European Union**. |
| Turkish | **Erdoğan**, **Türkiye**'de twitter yasaklandı. | **Erdogan** banned twitter in **Turkey**. |
| Arabic | قال غاريث بيل ، نجم فريق ريال مدريد ، إن الفوز بدوري أبطال أوروبا سيظل معه الى الأبد، وذلك بعدما ساعد هدفه النادي الملكي على هزيمة نادي اتليتكو مدريد في **Spain** . | Said **Gareth Bell**, the star of **Real Madrid**, said that winning the **Champions League** will remain with him "forever", and that after his goal helped the club to defeat **Atletico Madrid** in **Spain**. |
| Indonesian | **Rendjambe** meninggal dalam keadaan tidak jelas , yang mengakibatkan kerusuhan oleh pendukung oposisi marah di **Port Gentil** - dan **Libreville**. | **Rendjambe** died in unclear circumstances, which resulted in riots by angry opposition supporters in **Port Gentil** - and **Libreville**. |
| Chinese | 新华社 25 日报道 援引 新疆 公安厅 。 | **Xinhua News Agency** report quoted the 25th **Xinjiang** Public Security Department. |
| Greek | Ενλίλ έφερε τα Γκούτιους κάτω από το λόφους ανατολικά του Τίγρη, να φέρει το θάνατο σε ολόκληρη τη Μεσοποταμία. | **Enlil** brought the Gutians down from the hills east of the **Tigris**, to bring death throughout **Mesopotamia**. |

(a) Good Examples.

| Language | Sentence | Translation |
|---|---|---|
| English | He built **Spokane House** and **Kootanae** House and helped **David Thompson** cross the Continental Divide and discover the **Columbia** River. | - |
| Hungarian | Ezt követően, a csapat volt ismert, mint a **Tacoma** Vontató (1979), és még egyszer, a **Tacoma Tigers**. | Subsequently, the team was known as the **Tacoma** Tugs (1979) and, once more, the **Tacoma Tigers**. |
| Spanish | **DGG** también era dueño de **Polydor Records**. | **DGG** also owned **Polydor Records**. |
| Russian | Русский Федерация под руководством Владимира Путина в приложении Крым. | Russian **Federation** under the leadership of **Vladimir Putin** annexed **Crimea**. |
| Korean | 검찰은 유 병출 유엔, 서울 에서 운영 제주 해양 (주)의 소유자의 홈을 급습했다. | Prosecutors raided the home of **Yoo** Byung- **un**, the owner of **Jeju** Marine Co. Ltd, which operates in **Seoul**. |
| French | En 1970 , **Burgess** a été le candidat **républicain** succès pour le lieutenant - gouverneur et a servi deux mandats , de 1971 à 1975 . | In 1970, **Burgess** was the successful **Republican** candidate for Lieutenant - Governor and served two terms from 1971 to 1975. |
| Turkish | 1979 yılında , o folha sol ve kısa ömürlü Jornal da **República** ' da **Mino Carta** ile çalışmaya başladı . | In 1979, he left folha and short-lived Jornal da **República** began working with **Mino Carta**. |
| Arabic | وكان البابا يتحدث عن القدس وإلى جانبه الرئيس الفلسطيني محمود عباس بعد وصوله مباشرة إلى مدينة بيت لحم . | The **Pope** speaks of **Jerusalem** and to his part, **Palestinian** President **Mahmoud Abbas** after his arrival directly to the city of **Bethlehem**. |
| Indonesian | Ia lahir di Totowa, **New Jersey** dan meninggal di **Brooklyn**, **New York**. | He was born in Totowa, **New Jersey** and died in **Brooklyn**, **New York**. |
| Chinese | 周 先生 是 四川省 党委 书记 成为 中国 公安部 负责人 在2003 年 以前 。 | Mr. Zhou was the party secretary in **Sichuan province** before becoming head of China's Public Security Ministry in 2003. |
| Greek | Ήταν ο μόνος Αμερικανός για να χρησιμεύσει ως Πρέσβης στη Γαλλία, της Δημοκρατίας της Γερμανίας και το Ηνωμένο Βασίλειο. | He was the only **American** to serve as Ambassador to **France**, the **Republic of Germany** and the **United Kingdom**. |

(b) Bad examples

Table 5: POLYGLOT-NER results on several languages. Color code: {**Person**, **Location**, **Organization**}. We denote errors as the following: false positive, false negative, and label misclassification. Translations are acquired through Google Translate and labels on translated phrases correspond to their annotations in the source language.

and annotators' disagreements. Specifically, the English CoNLL dataset has an over representation of upper case words and sports teams. In the Dutch dataset, country names were abbreviated after the journalist name. For example, *Spain* will be mapped to *Spa* and *Italy* to *Ita*. This leads to more out of vocabulary (OOV) terms for which Polyglot does not have embeddings. Since we do not rely on any CoNLL-tailored preprocessing steps, such notational and stylistic differences affect our performance more than other approaches that tailor their systems to these differences at the cost of scalability. Such notational differences pose more harm to our performance than other approaches because we do not rely on CoNLL-tailored preprocessing steps.

However, [21] show that Wikipedia is better suited than human annotated datasets as a source of training for domain adaptation scenarios. We expect annotators trained on Wikipedia to be work better on heterogeneous content such as websites.

## 6 Distant Evaluation

Scarcity of human annotated datasets has limited the scope of our evaluation so far. We seek to use Statistical Machine Translation (SMT) as a tool to generate automated evaluation datasets. However, translation is not a one-to-one term mapping between two languages; the generated sentences may not preserve the word count or order. This poses a challenge in mapping the annotations from the source language to the target one. Therefore, we do not use the generated datasets for training, but rather for evaluation. We rely on comparing aggregated statistics over sentence translation pairs as an indicator of the quality of our annotators. We call this approach *Distant Evaluation* to emphasize the indirect connection between our comparative measures and the annotations quality. To simplify our approach, we assume:

- SMT is able to translate named entities from the source language to their corresponding phrases in the target language.
- SMT preserves the number of named entities mentioned.

We will show later these assumptions hold with varying degree across languages.

Specifically, we define the set of entity phrases appearing in a sentence ($S$) to be $\mathcal{P}$. Each phrase $p \in \mathcal{P}$ belongs to a category $T(p) \in \mathcal{Y}$. For each category $e \in \mathcal{Y}$, we define

$$C_e(S) = \sum_{p \in \mathcal{P}} [T(p) = e],$$

where $[x]$ is the indicator function. We define the sets of sentences that belong to the source language and the target language to be $L_1$, $L_2$, respectively. We define two classes of error measures: omitting entities $\mathcal{E}_{\mathcal{M}}$ and adding entities $\mathcal{E}_{\mathcal{A}}$, as the following

$$Z_e = \sum_{S \in L_1} C_e(S),$$
$$\mathcal{E}_{\mathcal{M}}(e) = \frac{1}{Z_e} \sum_{(S_1,S_2) \in (L_1,L_2)} |C_e(S_1) - C_e(S_2)|_+,$$
$$\mathcal{E}_{\mathcal{A}}(e) = \frac{1}{Z_e} \sum_{(S_1,S_2) \in (L_1,L_2)} |C_e(S_2) - C_e(S_1)|_+,$$

where $(S_1, S_2)$ is the sentence translation pair and $|x|_+ = max(0, x)$.

Next, we calculate $\mathcal{E}_{\mathcal{M}}$ and $\mathcal{E}_{\mathcal{A}}$ over all language pairs according the following steps:

- Annotate English Wikipedia sentences using STANFORD NER[5].
- Set $L_1$ to the above annotated sentences.
- Randomly pick 1500 sentences that have at least one entity detected.
- Translate these sentences using Google Translate to 40 languages.
- Calculate $\mathcal{E}_{\mathcal{M}}$ and $\mathcal{E}_{\mathcal{A}}$ for the language pairs for each entity type in $\mathcal{Y}^+$.

Figure 3 shows the performance of our system compared to other annotators; OPENNLP {English, Spanish, Dutch} [2], NLTK English [3], and STANFORD German [11]. Our new metrics are consistent our CoNLL evaluation, English outperforms both Spanish and Dutch. We outperform OPENNLP and NLTK, by a significant margin. POLYGLOT-NER German annotator covers more PERSON entities than STANFORD without adding many false positives. We notice that languages with the large number of Wikipedia articles like English (`en`), French (`fr`), Spanish (`es`) and Portuguese (`pt`) show strong performance. Moreover, our performance vary across categories, the annotators performing the best on PERSON category followed by LOCATION and then ORGANIZATION.

Our benchmark also highlights language specific issues. The poor performance in Japanese (`ja`) is due to the mismatch between the embedding vocabulary and the evaluation tokenizer. Vietnamese (`vi`) annotator aggressively annotates chunks as LOCATION ($\mathcal{E}_{\mathcal{A}} = 0.6$) because Vietnamese Freebase distribution of attributes is skewed towards LOCATION (See Figure 3b).

The quality of our metric is directly correlated to the quality of translations; while in general the above mentioned assumptions hold true for most translation pairs, we found some exceptions. First, Google Translate does not translate the entities efficiently in some languages, for example, "*이러한 이벤트는 Suriyothai 의 전설 , Chatrichalerm Yukol 감독 2001 태국 영화 에 묘사 된다.*" is the Korean translation of an English sentence *" These events are depicted in The Legend of Suriyothai, a 2001 Thai film directed by Chatrichalerm Yukol"*. This affects performance measure of Korean (`ko`), Greek (`el`) and Thai (`th`). Second, entity counts may not be preserved, for example, this Spanish translation *"Yehuda Magidovitch (1886-1961) fue uno de los arquitectos más prolíficos de Israel."* contains one location '`Israel`' that does not appear in the original English sentence *"Yehuda Magidovitch (1886–1961) was one of the most prolific Israeli architects."*

[5] http://nlp.stanford.edu/software/CRF-NER.shtml

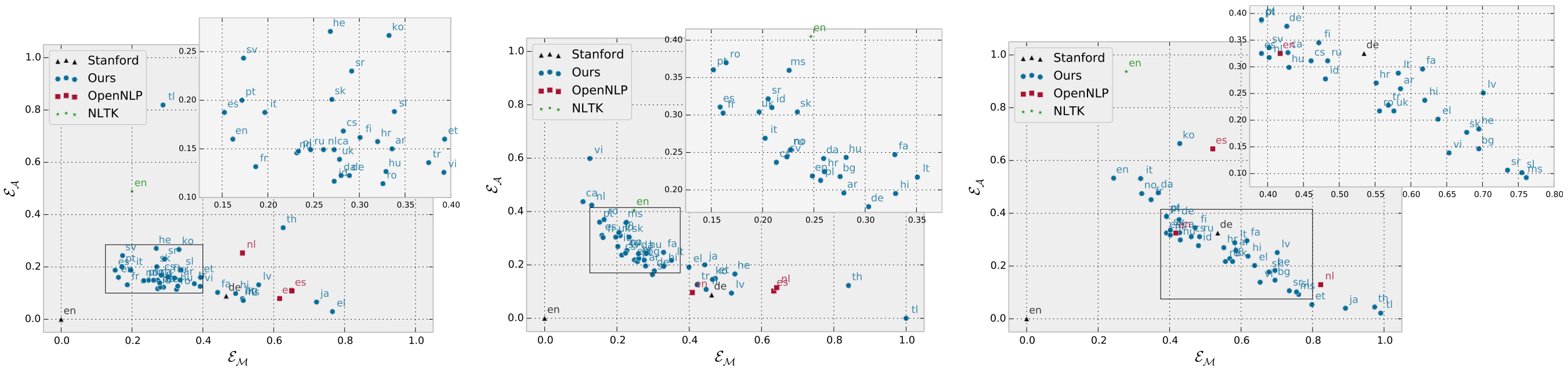

(a) PERSON (b) LOCATION (c) ORGANIZATION

Figure 3: Error Analysis. *Closer to the origin is better* STANFORD English NER is at the origin because it is the source of annotations.

We also investigate the effect of Wikipedia article counts on our performance. The size of Wikipedia of each language, affects our system in several aspects; Larger Wikipedia results in better word embeddings. Freebase has better attributes coverage for larger Wikipedias. More diverse set of training examples could be extracted from larger Wikipedia. Figure 4 shows the average error over all categories versus Wikipedia number of articles. We observe that larger Wikipedias result in many fewer false negatives with $\mathcal{E}_\mathcal{M}$ dropping by 0.6 (Figure 4a). On the other hand, larger Wikipedias annotates slightly more aggressively increasing $\mathcal{E}_\mathcal{A}$ by 0.15 (Figure 4b).

## 7 Conclusion & Future Work

We successfully built a multilingual NER system for 40 languages with no language specific knowledge or expertise. We use automatically learned features, and apply language agnostic data processing techniques. The system outperforms previous work in several languages and competitive in the rest on human annotated datasets. We demonstrate its performance on the rest of the languages, by a comparative analysis using machine translation. Our approach yields highly consistent performance across all languages. Wikipedia Cross-lingual links will be used in combination with Freebase to extend our approach to all languages as future work.

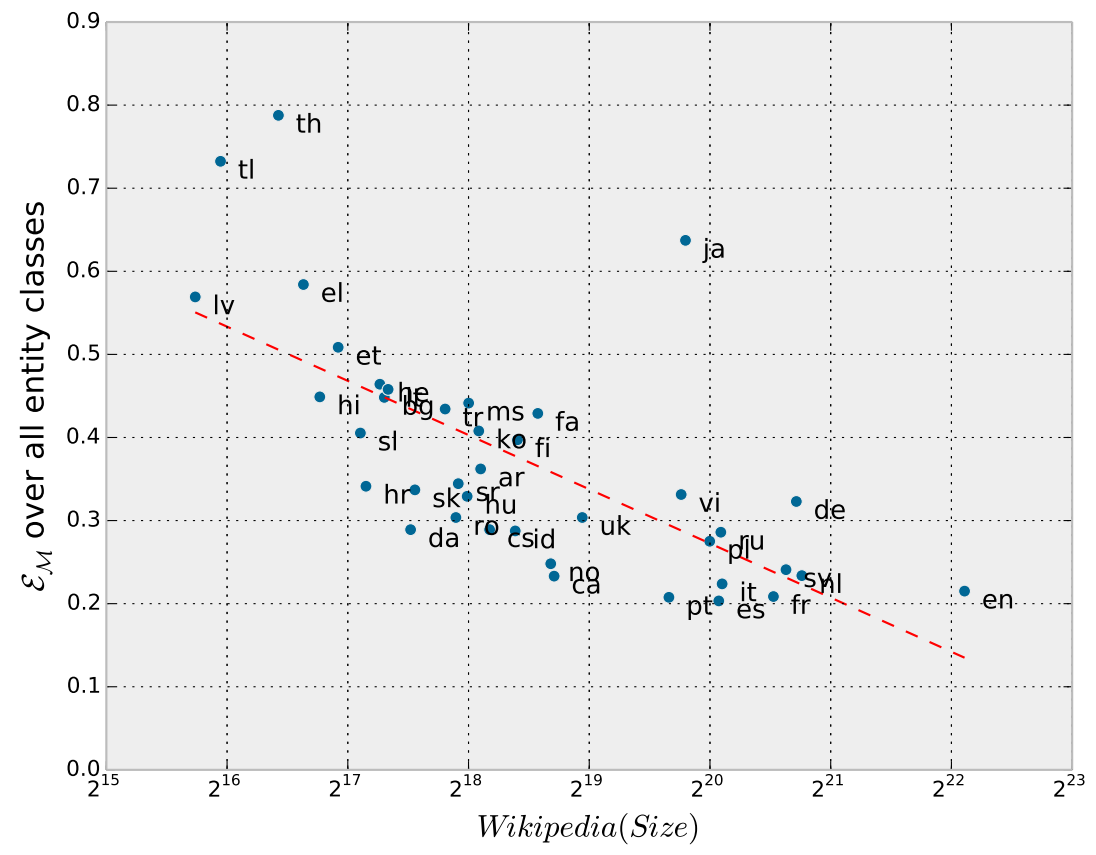

(a) $\mathcal{E}_\mathcal{M}$ of all entity classes versus Wikipedia article count.

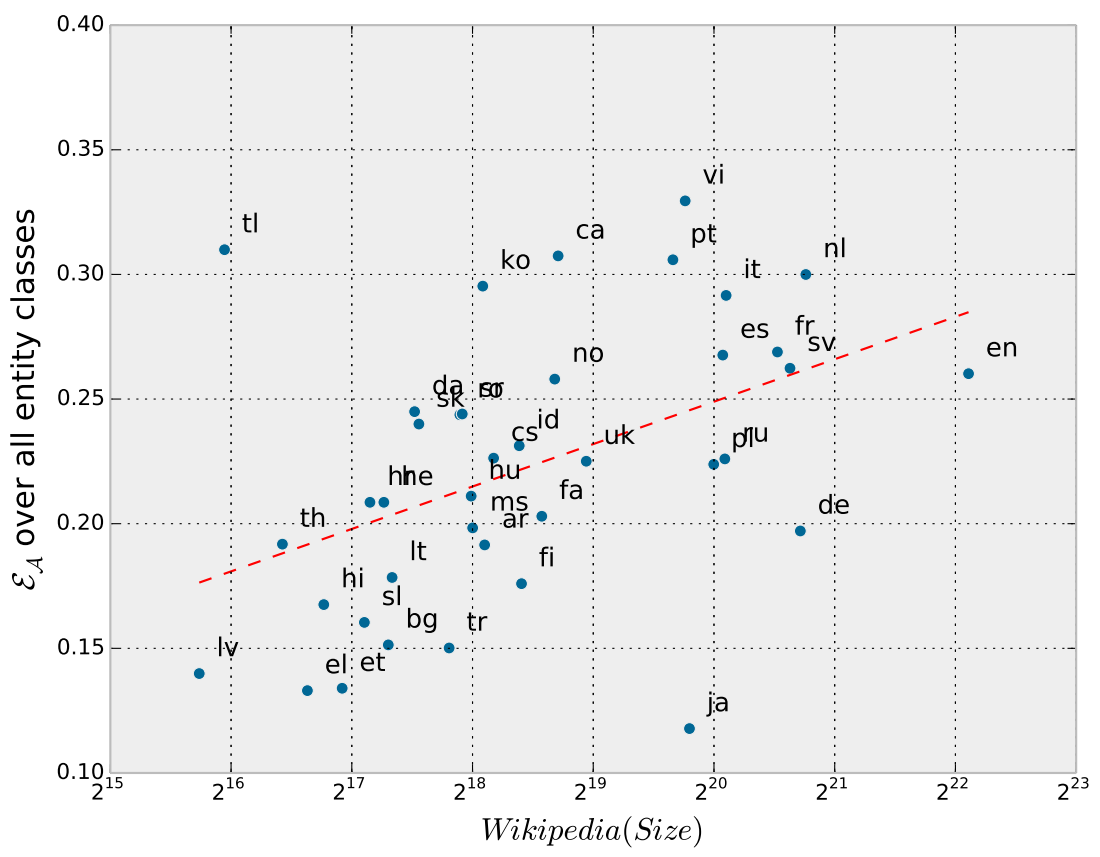

(b) $\mathcal{E}_\mathcal{A}$ of all entity classes versus Wikipedia article count.

Figure 4: Aggregated errors over all categories versus Wikipedia size for each language.